\documentclass[10pt,twocolumn,letterpaper]{article}

\usepackage{cvpr}
\usepackage{times}
\usepackage{epsfig}
\usepackage{graphicx}
\usepackage{amsmath}
\usepackage{amssymb}
\usepackage{bm}

\usepackage[breaklinks=true,colorlinks,bookmarks=false]{hyperref}

\cvprfinalcopy 

\def\cvprPaperID{****} 

\ifcvprfinal\fi
\begin{document}

\title{Deep Learning on Lie Groups for Skeleton-based Action Recognition}

\author{Zhiwu Huang$^\dagger$, Chengde Wan$^\dagger$, Thomas Probst$^\dagger$, Luc Van Gool$^{\dagger\ddagger}$\\
	$^\dagger$Computer Vision Lab, ETH Zurich, Switzerland \quad $^\ddagger$VISICS, KU Leuven, Belgium\\
	{\tt\small \{zhiwu.huang, wanc, probstt, vangool\}@vision.ee.ethz.ch}}

\maketitle

\begin{abstract}
  In recent years, skeleton-based action recognition has become a popular 3D classification problem. State-of-the-art methods typically first represent each motion sequence as a high-dimensional trajectory on a Lie group with an additional dynamic time warping, and then shallowly learn favorable Lie group features. In this paper we incorporate the Lie group structure into a deep network architecture to learn more appropriate Lie group features for 3D action recognition. Within the network structure, we design rotation mapping layers to transform the input Lie group features into desirable ones, which are aligned better in the temporal domain. To reduce the high feature  dimensionality, the architecture is equipped with rotation pooling layers for the elements on the Lie group. Furthermore, we propose a logarithm mapping layer to map the resulting manifold data into a tangent space that facilitates the application of regular output layers for the final classification. Evaluations of the proposed network for standard 3D human action recognition datasets clearly demonstrate its superiority over existing shallow Lie group feature learning methods as well as most conventional deep learning methods.
\end{abstract}

\section{Introduction}

Due to the development of depth sensors, 3D human activity analysis \cite{lv2006recognition,wang2012mining, hussein2013human, wang2013approach, vemulapalli2014human, anirudh2016elastic,vemulapallirolling,shahroudy2016ntu, wang2016mining,liu2016spatio,presti20163d, han2016space} has attracted more interest than ever before. Recent manifold-based approaches are quite successful at 3D human action recognition thanks to their view-invariant manifold-based representations for skeletal data. Typical examples include shape silhouettes in the Kendall's shape space \cite{veeraraghavan2005matching,anirudh2016elastic}, linear dynamical systems on the Grassmann manifold \cite{turaga2009locally}, histograms of oriented optical flow on a hyper-sphere \cite{chaudhry2009histograms}, and pairwise transformations of skeletal joints on a Lie group \cite{vemulapalli2014human, anirudh2016elastic, vemulapallirolling}. In this paper, we focus on studying manifold-based approaches \cite{vemulapalli2014human,anirudh2016elastic,vemulapallirolling} to learn more appropriate Lie group representations of skeletal action data, that have achieved state-of-the-art performances for some 3D human action recognition benchmarks.

As studied in \cite{vemulapalli2014human, anirudh2016elastic, vemulapallirolling}, Lie group feature learning methods often suffer from speed variations (i.e., temporal misalignment), which tend to deteriorate classification accuracy. To handle this issue, they typically employ dynamic time warping (DTW), as originally used in speech processing \cite{muller2007information}. Unfortunately, such process costs additional time, and also results in a two-step system that typically performs worse than an end-to-end learning scheme. Moreover, such Lie group representations for action recognition tend to be extremely high-dimensional, in part because the features are extracted per skeletal segment and then stacked. As a result, any computation on such nonlinear trajectories is expensive and complicated. To address this problem, \cite{vemulapalli2014human,anirudh2016elastic,vemulapallirolling} attempt to first flatten the underlying manifold via tangent approximation or rolling maps, and then exploit SVM or PCA-like method to learn features in the resulting flattened space. Although these methods achieve some success, they merely adopt shallow linear learning schemes, yielding sub-optimal solutions on the specific nonlinear manifolds.

Deep neural networks have shown their great power in learning compact and discriminative representations for images and videos, thanks to their ability to perform nonlinear computations and the effectiveness of gradient descent training with backpropagation. This has motivated us to build a deep neural network architecture for representation learning on Lie groups. In particular, inspired by the classical manifold learning theory \cite{tenenbaum2000global,roweis2000nonlinear,belkin2003laplacian,donoho2003hessian, huang2015log, huang2015projection}, we equip the new network structure with rotation mapping layers, with which the input Lie group features are transformed to new ones with better alignment. As a result, the effect of speed variations can be appropriately mitigated. In order to reduce the high dimensionality of the Lie group features, we design special pooling layers to compose them in terms of spatial and temporal levels, respectively. As the output data reside on nonlinear manifolds, we also propose a Riemannian computing layer, whose outputs could be fed into any regular output layers such as a softmax layer. In short, our main contributions are:
\begin{itemize}
	\item A novel neural network architecture is introduced to deeply learn more desirable Lie group representations for the problem of skeleton-based action recognition.
	\item The proposed network provides a paradigm to incorporate the Lie group structure into deep learning, which generalizes the traditional neural network model to non-Euclidean Lie groups.
	\item To train the network within the backpropagation framework, a variant of stochastic gradient descent optimization is exploited in the context of Lie groups.
\end{itemize}

\section{Relevant Work}

Already quite some works \cite{zbikowski1993lie,pearson1995hopfield,albertini1992neural,moreau1996lie,pearson1995changing, fiori2002unsupervised, gens2014deep} have applied aspects of Lie group theory to deep neural networks. For example, \cite{pearson1995changing} investigated how stability properties of a continuous recursive neural network can be altered within neighbourhoods of equilibrium points by the use of Lie group projections operating on the synaptic weight matrix. \cite{fiori2002unsupervised} studied the behavior of unsupervised neural networks with orthonormality constraints, by exploiting the differential geometry of Lie groups. In particular, two sub-classes of the general Lie group learning theories were studied in detail, tackling first-order (gradient-based) and second-order (non-gradient-based) learning. \cite{gens2014deep} introduced deep symmetry networks (symnets), a generalization of convolutional networks that forms feature maps over arbitrary symmetry groups that are basically Lie groups. The symnets utilize kernel-based interpolation to tractably tie parameters and pool over symmetry spaces of any dimension.

Moreover, recently some deep learning models have emerged \cite{bruna2014spectral,boscaini2015learning,masci2015geodesic,jain2016structural,huang2016riemannian,huang2016building} that deal with data in a non-Euclidean domain. For instance, \cite{bruna2014spectral} proposed a spectral version of convolutional networks to handle graphs. It exploits the notion of non shift-invariant convolution, relying on the analogy between the classical Fourier transform and the Laplace-Beltrami eigenbasis. \cite{jain2016structural} developed a scalable method for treating an arbitrary spatio-temporal graph as a rich recurrent neural network mixture, which can be used to transform any spatio-temporal graph by employing a certain set of well-defined steps. For shape analysis, \cite{masci2015geodesic} proposed a `geodesic convolution' on local geodesic coordinate systems to extract local patches on the shape manifold. This approach performs convolutions by sliding a window over the manifold, and local geodesic coordinates are used instead of image patches. To deeply learn symmetric positive definite (SPD) matrices - used in many tasks - \cite{huang2016riemannian} developed a Riemannian network on the manifolds of SPD matrices, with some layers specially designed to deal with such structured matrices. 

In summary, such works have applied some theories of Lie groups to regular networks, and even generalized the common networks to non-Euclidean domains. Nevertheless, to the best of our knowledge, this is the first work that studies a deep learning architecture on Lie groups to handle the problem of skeleton-based action recognition.

\begin{figure*}[t]
	\begin{center}
		\includegraphics[width=0.85\linewidth]{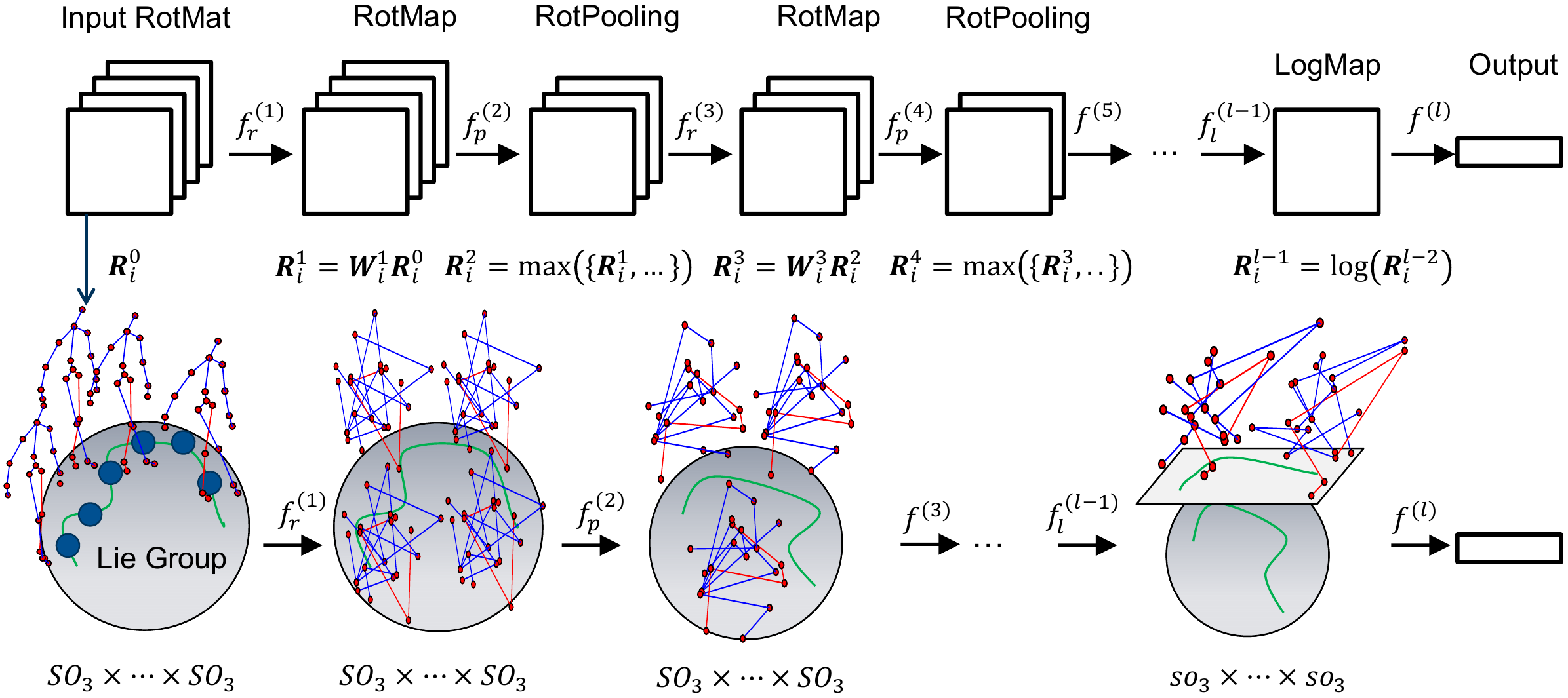}
	\end{center}
	\caption{Conceptual illustration of the proposed Lie group Network (LieNet) architecture. In the network structure, the data space of each RotMap/RotPooling layer corresponds to a Lie group, while the weight spaces of the RotMap layers are Lie groups as well.}
	\label{fig:long}
	\label{Fig1}
\end{figure*}


\section{Lie Group Representation for Skeletal Data} \label{sec3}


Let $S=(V,E)$ be a body skeleton, where $V=\{v_1,\ldots,v_N\}$ denotes the set of body joints, and $E=\{\bm{e}_1,\ldots,\bm{e}_M\}$ indicates the set of edges, i.e. oriented rigid body bones. As studied in \cite{vemulapalli2014human,anirudh2016elastic,vemulapallirolling}, the relative geometry of a pair of body parts $\bm{e}_n$ and $\bm{e}_m$ can be represented in a local coordinate system attached to the other. The local coordinate system of body part $\bm{e}_n$ is calculated by rotating with minimum rotation so that its stating joint
becomes the origin and it coincides with the $x$-axis. With the process, we consequently get the transformed 3D vectors $\hat{\bm{e}}_m, \hat{\bm{e}}_n$ for the two edges $\bm{e}_m, \bm{e}_n$ respectively. Then we can compute the rotation matrix $\bm{R}_{m,n}$ ($\bm{R}_{m,n}^T \bm{R}_{m,n}=\bm{R}_{m,n} \bm{R}_{m,n}^T = I_n, |\bm{R}_{m,n}| = 1$) from $\bm{e}_m$ to the local coordinate system of $\bm{e}_n$. Specifically, we can firstly calculate the axis-angle representation $(\bm{\omega}, \theta)$ for the rotation matrix $\bm{R}_{m,n}$  by
\begin{alignat}{2}
	\bm{\omega} & = \frac{\hat{\bm{e}}_m  \otimes \hat{\bm{e}}_n}{\|\hat{\bm{e}}_m \otimes \hat{\bm{e}}_n\|}, \label{Eq0.0} \\
	\theta & = \arccos(\hat{\bm{e}}_m \cdot \hat{\bm{e}}_n). \label{Eq0.1}
\end{alignat}
where $\otimes, \cdot$ are outer and inner products respectively. Then, the axis-angle representation can be easily transformed to a rotation matrix $\bm{R}_{m,n}$.
In the same way, the rotation matrix $\bm{R}_{n,m}$ from $\bm{e}_n$ to the local coordinate system of $\bm{e}_m$ can be computed. To fully encode the relative geometry between $\bm{e}_m$ and $\bm{e}_n$, $\bm{R}_{m,n}$ and $\bm{R}_{n,m}$ are both used. As a result, a skeleton $S$ at the time instance $t$ is represented by the form $(\bm{R}_{1,2}(t), \bm{R}_{2,1}(t)\ldots, \bm{R}_{M-1,M}(t), \bm{R}_{M,M-1}(t))$, where $M$ is the number of body parts, and the number of rotation matrices is $2C_{M}^{2}$ ($C_{M}^{2}$ is the combination formula).

The set of $n\times n$ rotation matrices in $\mathbb{R}^n$ forms the special orthogonal group $SO_n$ which is actually a matrix Lie group \cite{huper2007geometry,boumal2011discrete,hall2015lie}. Accordingly, each motion sequence of a moving skeleton is represented with a curve on the Lie group $SO_3 \times \ldots \times SO_3$. It is known that the matrix Lie group is endowed with a Riemannian manifold structure that is differentiable. Hence, at each point $\bm{R}_0$ on $SO_n$, one can derive the tangent space $T_{\bm{R}_0}SO_n$ that is a vector space spanned by the set of skew-symmetric matrices. When the anchor point is the identity matrix $\bm{I}_n \in SO_n$, the resulting tangent space is known as the Lie algebra $so_n$. As the tangent spaces are equipped with the inner product, the Riemannian metric on $SO_n$ can be defined by the Frobenius inner product:
\begin{equation}
	<\bm{A}_1,\bm{A}_2>=trace(\bm{A}_1^T\bm{A}_2), \bm{A}_1, \bm{A}_2 \in T_{\bm{R}_0}SO_n.
	\label{Eq1}
\end{equation}

The logarithm map $\log_{\bm{R}_0}$ and exponential map $\exp_{\bm{R}_0}$ at $\bm{R}_0$ on $SO_n$ associated with the Riemannian metric can be expressed in terms of the usual matrix logarithm $\log$ and exponential $\exp$ as
\begin{alignat}{2}
	\log_{\bm{R}_0}(\bm{R}_1)&=\log(\bm{R}_1\bm{R}_0^T) 
	\hspace{0.2cm} \mbox{with} \hspace{0.2cm} 
	\bm{R}_0, \bm{R}_1 \in SO_n, \label{Eq2} 
	\\
	\exp_{\bm{R}_0}(\bm{A}_1)&=\exp^{\bm{A}_1\bm{R}_0^T} 
	\hspace{0.2cm} \mbox{with} \hspace{0.2cm} 
	\bm{A}_1 \in T_{\bm{R}_0}SO_n. \label{Eq3}
\end{alignat}

\section{Lie Group Network for Skeleton-based Action Recognition}

For the problem of skeleton-based action recognition, we build a deep network architecture to learn the Lie group representations of skeletal data. The network structure is dubbed as LieNet, where each input is an element on the Lie Group. Like convolutional networks (ConvNets), the LieNet also exhibits fully connected convolution-like layers and pooling layers, named rotation mapping (RotMap) layers and rotation pooling (RotPooling) layers respectively. In particular, the proposed RotMap layers perform transformations on input rotation matrices to generate new rotation matrices, which have the same manifold property, and are expected to be aligned more accurately for more reliable matching. The RotPooling layers aim to pool the resulting rotation matrices at both spatial and temporal levels such that the Lie group feature dimensionality can be reduced. Since the rotation matrices reside on non-Euclidean manifolds, we have to design a layer named logarithm mapping (LogMap) layer, to perform the Riemannian computations on. This transforms the rotation matrices into the usual skew-symmetric matrices, which lie in Euclidean space and hence can be fed into any regular output layers. The architecture of the proposed LieNet is shown in Fig.\ref{Fig1}.

\subsection{RotMap Layer}

As well-known from classical manifold learning theory \cite{tenenbaum2000global,roweis2000nonlinear,belkin2003laplacian,donoho2003hessian, huang2015log, huang2015projection}, one can learn or preserve the original data structure to faithfully maintain geodesic distances for better classification. Accordingly, we design a RotMap layer to transform the input rotation matrices to new ones that are more suitable for the final classification. Formally, the RotMap layers adopt a rotation mapping $f_r$ as
\begin{equation}
	\begin{aligned}
		& f_r^{(k)}((\bm{R}^{k-1}_1, \bm{R}^{k-1}_2 \ldots, \bm{R}^{k-1}_{\hat{M}}); \bm{W}^k_1,\bm{W}^k_2 \ldots, \bm{W}^k_{\hat{M}}) \\
		& = (\bm{W}^k_1 \bm{R}^{k-1}_1, \bm{W}^k_2 \bm{R}^{k-1}_2 \ldots, \bm{W}^k_{\hat{M}} \bm{R}^{k-1}_{\hat{M}})
		\\& = (\bm{R}^k_1, \bm{R}^k_2 \ldots, \bm{R}^k_{\hat{M}})
		\label{Eq4}
	\end{aligned}
\end{equation}
where $\hat{M}=2C_{M}^{2}$ ($M$ is the number of body bones in one skeleton, $C_{M}^{2}$ is the combination computation),  $(\bm{R}^{k-1}_1, \bm{R}^{k-1}_2 \ldots, \bm{R}^{k-1}_{\hat{M}}) \in {SO}_3 \times {SO}_3 \ldots \times {SO}_3$ is the input Lie group feature (i.e., product of rotation matrices) for one skeleton in the $k$-th layer, $\bm{W}^k_i  \in  \mathbb{R}^{3 \times 3}$ is the transformation matrix (connection weights), and $(\bm{R}^k_1, \bm{R}^k_2 \ldots, \bm{R}^k_{\hat{M}})
$ is the resulting Lie group representation. Note that although there is only one transformation matrix for each rotation matrix, it would be easily extended with multiple projections for each input. To ensure the form $(\bm{R}^k_1, \bm{R}^k_2 \ldots, \bm{R}^k_{\hat{M}})$ becomes a valid product of rotation matrices residing on ${SO}_3 \times {SO}_3 \ldots \times {SO}_3$, the transformation matrices $\bm{W}^k_1,\bm{W}^k_2,\ldots, \bm{W}^k_{\hat{M}}$ are all basically required to be rotation matrices. Accordingly, both the data and the weight spaces on each RotMap layer correspond to a Lie group ${SO}_3 \times {SO}_3 \ldots \times {SO}_3$. 

Since the RotMap layers are designed to work together with the classification layer, each resulting skeleton representation is tuned for more accurate classification in an end-to-end deep learning manner. In other words, the major purpose of designing the RotMap layers is to align the Lie group representations of a moving skeleton for more faithful matching.


\subsection{RotPooling Layer}

In order to reduce the complexity of deep models, it is typically useful to reduce the size of the representations to decrease the amount of parameters and computation in the network.
For this purpose, it is common to insert a pooling layer in-between successive convolutional layers in a typical ConvNet architecture. The pooling layers are often designed to compute statistics in local neighborhoods, such as sum aggregation, average energy and maximum activation.



\begin{figure}[t]
	\begin{center}
		\includegraphics[width=0.95\linewidth]{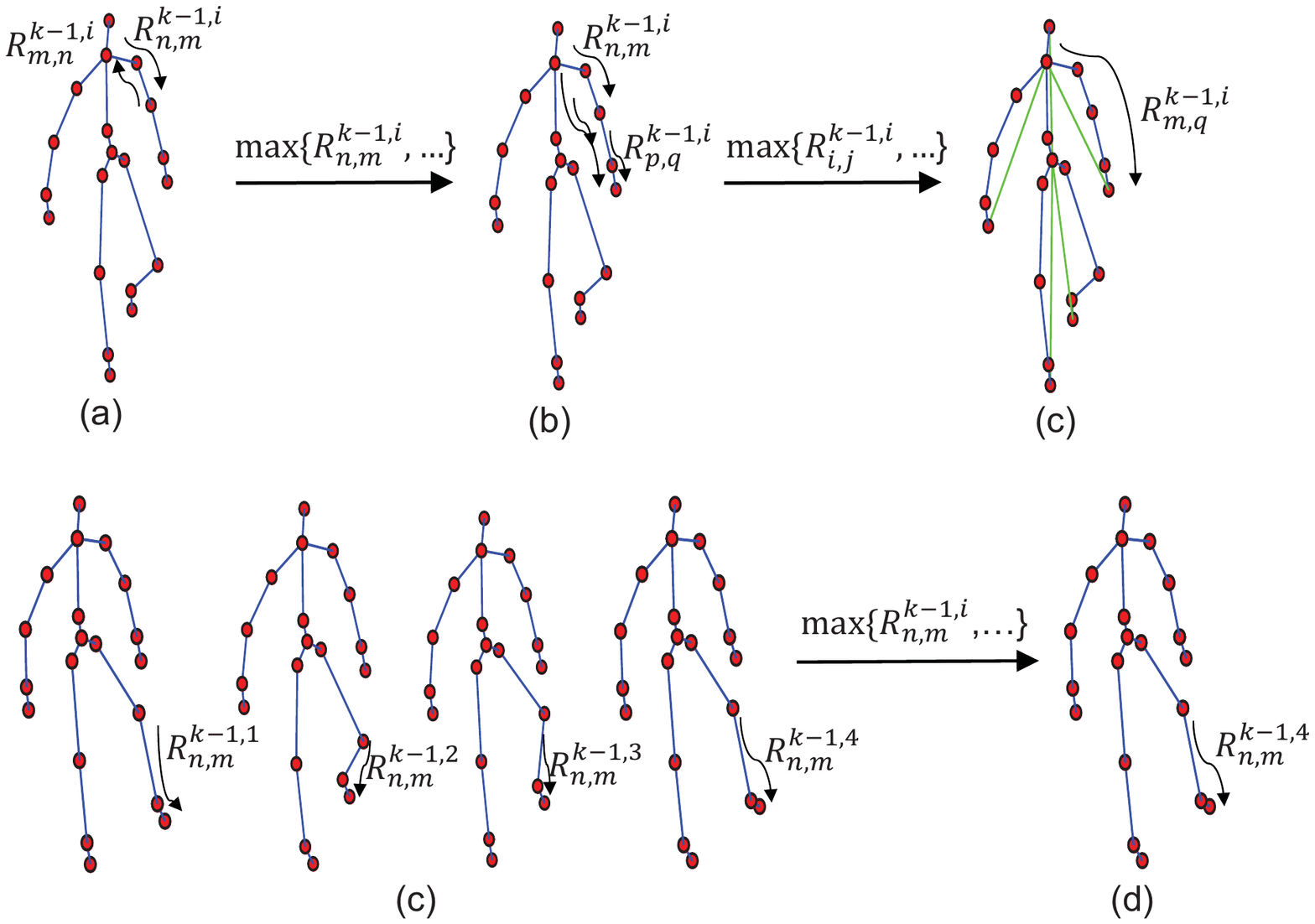}
	\end{center}
	\caption{Illustration of spatial pooling (SpaPooling) (a)$\rightarrow$(b)$\rightarrow$(c) and temporal pooling (TemPooling)  (c)$\rightarrow$(d) schemes.}
	\label{fig:long}
	\label{Fig2}
\end{figure}

Without loss of generality, we here just introduce max pooling\footnote{In contrast to sum and mean poolings, max pooling can generate valid rotation matrices directly, and hence suits the proposed LieNets. On the other hand, leveraging Lie group computing to enable sum and mean pooling to work for the LieNets, however, goes beyond the scope of this paper.} to the LieNet setting with the equivalent notion of neighborhood. 
Since the input and output of the special pooling layers are both expected to be rotation matrices, we call this kind of layers as rotation pooling (RotPooling) layers. For the RotPooling, we propose two different concepts of neighborhood in this work. The first one is on the spatial level. As shown in Fig.\ref{Fig2}(a)$\rightarrow$(b), we first pool the Lie group features on each pair of basic bones $\bm{e}_m, \bm{e}_n$ in the $i$-th frame, which is represented by the two rotation matrices $\bm{R}^{k-1,i}_{m,n}, \bm{R}^{k-1,i}_{n,m}$ (here $k-1$ is the order of the layer) as aforementioned. Then, as depicted in Fig.\ref{Fig2}(b)$\rightarrow$(c), we can perform pooling on the adjacent bones that belong to the same group (here, we can define five part groups, i.e., torso, two arms and two legs, of the body). However, the second step would inevitably result in a serious spatial misalignment problem, and thus lead to bad matching performances. Therefore, we finally only adopt the first step pooling.
In this setting, the function of the max pooling is given by
\begin{equation}
	\begin{aligned}
		& f_p^{(k)}(\{\bm{R}^{k-1,i}_{m,n}, \bm{R}^{k-1,i}_{n,m}\}) =  \max(\{\bm{R}^{k-1,i}_{m,n}, \bm{R}^{k-1,i}_{n,m}\}) \\
		& = \begin{cases}
			\bm{R}^{k-1,i}_{m,n}, & \text{if} \quad \Theta(\bm{R}^{k-1,i}_{m,n}) > \Theta(\bm{R}^{k-1,i}_{n,m}) ,\\
			\bm{R}^{k-1,i}_{n,m}, & \text{otherwise},
		\end{cases}
		\label{Eq5}
	\end{aligned}
\end{equation}
where $\Theta(\cdot)$ is the representation of the given rotation matrix such as quaternion, Euler angle or Euler axis-angle.
For example, the Euler axis $\bm{\omega}$ and angle $\theta$ representations are typically calculated by
\begin{alignat}{2}
	\bm{\omega}(\bm{R}_{n,m})& = \frac{1}{2\sin(\theta(\bm{R}_{n,m}))} \begin{pmatrix}
		\bm{R}_{n,m}(3,2)-\bm{R}_{n,m}(2,3) \\
		\bm{R}_{n,m}(1,3)-\bm{R}_{n,m}(3,1) \\
		\bm{R}_{n,m}(2,1)-\bm{R}_{n,m}(1,2)
	\end{pmatrix}, \label{Eq7} \\
	\theta(\bm{R}_{n,m})&=\arccos \left(\frac{trace({\bm{R}_{n,m})-1}}{2}\right), \label{Eq6}
\end{alignat}
where $\bm{R}_{n,m}(i,j)$ is the $i$-the row, $j$-th column element of $\bm{R}_{n,m}$. 
Unfortunately, except the angle representation, it is non-trivial to define an ordering relation for a quaternion or an axis-angle representation. Hence, in this paper, we finally adopt the angle form Eqn.\ref{Eq6} of rotation matrices and its simple ordering relation to calculate the function $\Theta(\cdot)$.

The other pooling scheme is on the temporal level. As shown in Fig.\ref{Fig2} (c)$\rightarrow$(d), the aim of the temporal pooling is to obtain more compact representations for a motion sequence. This is because a sequence often contains many frames, which results in the problem of extremely high-dimensional representations. Thus, pooling in the temporal domain can reduce the model complexity as well. Formally, the function of this kind of max pooling is defined as
\begin{equation}
	\begin{aligned}
		& f_p^{(k)}(\{(\bm{R}^{k-1,1}_{1,2} \ldots \bm{R}^{k-1,1}_{M-1,M}) \ldots, (\bm{R}^{k-1,p}_{1,2}\ldots,\bm{R}^{k-1,p}_{M-1,M})\}) \\
		& = (\max(\{\bm{R}^{k-1, 1}_{1,2}\ldots, \bm{R}^{k-1,p}_{1,2}\})  \ldots, \\ & \quad\quad \max(\{\bm{R}^{k-1,1}_{M-1,M}\ldots, \bm{R}^{k-1, p}_{M-1,M}\})), \label{Eq9}
	\end{aligned}
\end{equation}
where $M$ is the number of body parts in one skeleton, $p$ is the number of skeleton frames for pooling, and the function $\max(\cdot)$ is defined in the way of Eqn.\ref{Eq5}.

\subsection{LogMap Layer}

Classification of curves on the Lie group  $SO_3 \times \ldots \times SO_3$ is a complicated task due to the non-Euclidean nature of the underlying space. To address the problem as in \cite{vemulapallirolling}, we design the logarithm map (LogMap) layer to flatten the Lie group $SO_3 \times \ldots \times SO_3$ to its Lie algebra $so_3 \times \ldots \times so_3$. Accordingly, by using the logarithm map Eqn.\ref{Eq2}, the function of this layer can be defined as
\begin{equation}
	\begin{aligned}
		& f_l^{(k)}((\bm{R}^{k-1}_1, \bm{R}^{k-1}_2 \ldots, \bm{R}^{k-1}_{\hat{M}})) \\
		& =(\log(\bm{R}^{k-1}_1), \log(\bm{R}^{k-1}_2) \ldots, \log(\bm{R}^{k-1}_{\hat{M}})). \label{Eq10}
	\end{aligned}
\end{equation}

One typical approach to calculate the logarithm map is to use the approach $\log(\bm{R})= \bm{U} \log(\bm{\Sigma}) \bm{U}^T$, where $\bm{R}=\bm{U}\bm{\Sigma}\bm{U}^T$, $\log(\bm{\Sigma})$ is the diagonal matrix of the eigenvalue logarithms. However, the spectral operation not only suffers from the problem of zeroes occurring in $\log(\Sigma)$ due to the property of the rotation matrix $\bm{R}$, but also consumes too much time for matrix gradient computation~\cite{ionescu2015matrix}. Therefore, we resort to other approaches to perform the function of this layer. Fortunately, we can explore the relationship between the logarithm map and the axis-angle representation as:
\begin{alignat}{1}
	\log(\bm{R})=
	\begin{cases}
		0, &  \text{if} \quad \theta(\bm{R}) =0,\\
		\frac{\theta(\bm{R})}{2\sin(\theta(\bm{R}))}(\bm{R}-\bm{R}^T), & \text{otherwise},
	\end{cases} \label{Eq11}
\end{alignat}
where $\theta(\bm{R})$ is the angle Eqn.\ref{Eq6} of $\bm{R}$. With this equation, the corresponding matrix gradient can be easily derived by traditional element-wise matrix calculation.

\subsection{Output Layers}

After performing the LogMap layers, the outputs can be transformed into vector form and concatenated directly frame by frame within one sequence due to their Euclidean nature. Then, we can add any regular network layers such as rectified linear unit (ReLU) layers and regular fully connected (FC) layers. In particular for the ReLU layer, we can simply set relatively small elements to zero as done in classical ReLU. In the FC layer, the dimensionality of the weight is set to ${d_{k} \times d_{k-1}}$, where $d_{k}$ and $d_{k-1}$ are the class number and the vector dimensionalities, respectively. For skeleton-based action recognition, we employ a common softmax layer as the final output layer. Besides, as studied in \cite{shahroudy2016ntu,liu2016spatio}, learning temporal dependencies over the sequential data can improve human action recognition. Hence, we can also feed the outputs into Long Short-Term Memory (LSTM) unit to learn useful temporal features. Because of the space limitation, we do not study this any further.


\section{Training Procedure}
\label{backprop}

In order to train the proposed LieNets, we exploit the Stochastic gradient descent (SGD) algorithm that is one of the most popular network training tools. To begin with, let the LieNet model be represented as a sequence of function compositions $f= f^{(l)} \circ f^{(l-1)} \ldots \circ f^{(1)}$ with a parameter tuple $\bm{W} = (\bm{W}_l, \bm{W}_{l-1} \ldots, \bm{W}_1)$, where $f^{(k)}$ is the function for the $k$-th layer, $\bm{W}_k$ (dropping the sample index for simplicity) represents the weight parameters of the $k$-th layer, and $l$ is the number of layers. The loss of the $k$-th layer is defined by $L^{(k)}=\ell \circ f^{(l)} \ldots  \circ f^{(k)}$, where $\ell$ is the loss function for the final output layer.

To optimize the deep model, one classical SGD algorithm needs to compute the gradient of the objective function, which is typically achieved by the backpropagation chain rule. In particular, the gradients of the weight $\bm{W}_k$ and the data $\bm{R}_{k-1}$ (dropping the sample index for simplicity) for the $k$-th layer can be respectively computed by the chain rule:
\begin{alignat}{2}
	\frac{\partial L^{(k)}(\bm{R}_{k-1}, y)}{\partial \bm{W}_k} &  = \frac{\partial L^{(k+1)}(\bm{R}_k, y)}{\partial \bm{R}_k}\frac{ \partial f^{(k)}(\bm{R}_{k-1})}{\partial \bm{W}_k}, \label{Eq12} \\
	\frac{\partial L^{(k)}(\bm{R}_{k-1}, y)}{\partial \bm{R}_{k-1}}  &= \frac{\partial L^{(k+1)}(\bm{R}_k, y)}{\partial \bm{R}_k}\frac{ \partial f^{(k)}(\bm{R}_{k-1})}{\partial \bm{R}_{k-1}}, \label{Eq13}
\end{alignat}
where $y$ is the class label, $\bm{R}_k=f^{(k)}(\bm{R}_{k-1})$. Eqn.\ref{Eq12} is the gradient for updating $\bm{W}_k$, while Eqn.\ref{Eq13} computes the gradients in the layers below to update $\bm{R}_{k-1}$.

The gradients of the data involved in RotPooling, LogMap and regular output layers can be calculated by Eqn.\ref{Eq13} as usual. Particularly, the gradient for the data in RotPooling can be computed with the same gradient computing approach used in a regular max pooling layer in the context of traditional ConvNets. For the data in the LogMap layer, the gradient can be obtained by the element-wise gradient computation on the involved rotation matrices.


On the other hand, the computation of the gradients of the parameter weights defined in the RotMap layers is non-trivial. This is because the weight matrices are enforced to be on the Riemannian manifold ${SO}_3$ of the rotation matrices, i.e. the Lie group. As a consequence, merely using Eqn.\ref{Eq12} to compute their Euclidean gradients rather than Riemannian gradients in the procedure of backpropagation would not generate valid rotation weights.
To handle this problem, we propose a new approach of updating the weights used in Eqn.\ref{Eq4} for the RotMap layers.  As studied in \cite{absil2008optimization}, the steepest descent direction for the used loss function $L^{(k)}(\bm{R}_{k-1}, y)$ with respect to $\bm{W}_k$ on the manifold ${SO}_3$ is the Riemannian gradient $\tilde{\nabla} L^{(k)}_{\bm{W}_k}$, which can be obtained by parallel transporting the Euclidean gradients onto the corresponding tangent space. In particular, transporting the gradient from a point $\bm{W}_k^{t}$ to another point $\bm{W}_k^{t+1}$ requires subtracting the normal component $\bar{\nabla} L^{(k)}_{\bm{W}_k}, \label{Eq14} $ at $\bm{W}_k^{t+1}$, which can be obtained as follows:
\begin{equation}
	\bar{\nabla} L^{(k)}_{\bm{W}_k}=\nabla L^{(k)}_{\bm{W}_k}\bm{W}_k^T\bm{W}_k,
	\label{Eq14}
\end{equation}
where the Euclidean gradient $\nabla L^{(k)}_{\bm{W}_k}$ is computed by using Eqn.\ref{Eq12} as
\begin{equation}
	\nabla L^{(k)}_{\bm{W}_k}=\frac{\partial L^{(k+1)}(\bm{R}_k, y)}{\partial \bm{R}_k}\bm{R}_{k-1}^T.
	\label{Eq17}
\end{equation}
Thanks to the parallel transport, the Riemannian gradient can be calculated by
\begin{equation}
	\tilde{\nabla} L^{(k)}_{\bm{W}_k} =\nabla L^{(k)}_{\bm{W}_k}-\bar{\nabla} L^{(k)}_{\bm{W}_k}.
	\label{Eq15}
\end{equation}

Searching along the tangential direction takes the update in the tangent space of the ${SO}_3$ manifold. Then, such update is mapped back to the ${SO}_3$ manifold with a retraction operation. Consequently, an update of the weight $\bm{W}_k$ on the ${SO}_3$ manifold is of the following form
\begin{alignat}{1}
	\bm{W}_k^{t+1} = \Gamma(\bm{W}_k^{t}-\lambda\tilde{\nabla} L^{(k)}_{\bm{W}_k}), \label{Eq16}
\end{alignat}
where $\bm{W}_k^{t}$ is the current weight, $\Gamma$ is the retraction operation, $\lambda$ is the learning rate.


\section{Experiments}

We employ three standard 3D human action datasets to study the effectiveness of the proposed LieNets.


\subsection{Evaluation Datasets}

\setlength{\parskip}{0.1\baselineskip}

\noindent \textbf{G3D-Gaming dataset \cite{bloom2012g3d}} contains 663 sequences of 20 different gaming motions. Each subject performed every action more than two times. Besides, 3D locations of 20 joints (i.e., 19 bones) are provided with the dataset.


\noindent \textbf{HDM05 dataset \cite{muller2007hdm}} consists of 2,337 sequences of 130 action classes executed by various actors. Most of the motion sequences have been performed several times by all five actors according to the guidelines in a script. As G3D-Gaming \cite{bloom2012g3d} dataset, 3D locations of 31 joints (i.e., 30 bones) of the subjects are provided as well with this dataset.

\noindent \textbf{NTU RGB+D dataset \cite{shahroudy2016ntu}} is, to the best of our knowledge, currently the largest 3D action recognition dataset, which contains more than 56,000 sequences. A total of 60 different action classes are performed by 40 subjects. 3D coordinates of 25 joints (i.e., 24 bones) are also offered. Due to its large scale, the dataset is highly suitable for deep learning.


\subsection{Implementation Details}

For the feature extraction, we use the code of~\cite{vemulapallirolling} to represent each human skeleton with a point on the Lie group $SO_3 \times \ldots \times SO_3$. As preprocessed in \cite{vemulapallirolling}, we normalize any sequence of motion into a fixed $N$-length one. As a result, for each moving skeleton, we finally compute a Lie group curve of length 100, 16, 64 for the G3D-Gaming, HDM05 and NTU RGB-D datasets, respectively.


As the focus of this work is on skeleton-based action recognition, we mainly utilize manifold-based approaches for comparison. The two baseline approaches are the special Euclidean group (SE) \cite{vemulapalli2014human} and the special orthogonal group (SO) \cite{vemulapallirolling} representations based shallow learning methods. For a fair comparison, we use the source codes from the original authors, and set the involved parameters as in the original papers. For the proposed LieNet, we build its architecture with single or multiple block(s) of RotMap/RotPooling layers illustrated in Fig.\ref{Fig1} before the three final layers being LogMap, FC and softmax layers. The learning rate $\lambda$ is fixed to 0.01, the batch size is set to 30, the weights in the RotMap layers are initialized as random rotation matrices, the number of samples for the temporal RotPooling (TemPooling) layer is set to 4. For training the LieNet, we just use an i7-6700K (4.00GHz) PC without any GPUs. As the LieNet gets promising results on all datasets with the same configuration, this shows its insensitivity to the parameter settings. Note that, for the LieNets, we do not employ the dynamic time warping (DTW) technique \cite{muller2007information}, which has been used in the SO and SE methods to solve the problem of speed variations.

\subsection{Experimental Results}

\noindent \textbf{G3D-Gaming dataset \cite{bloom2012g3d}}. For the dataset, we follow a cross-subject test setting, where half the subjects are used for training and the other half are employed for testing. All the results reported for this dataset are averaged over ten different combinations of training and testing datasets.

Table \ref{tab1} compares the proposed LieNet with the state-of-the-art methods (i.e., RBM-HMM \cite{nie2014capturing}, SE \cite{vemulapalli2014human} and SO \cite{vemulapallirolling}) reported for the G3D-Gaming dataset. To be fair, we report their results without using the Fourier Temporal Pyramid (FTP) post-processing (their accuracies are 91.09\% and 90.94\% after using FTP). As shown in Table \ref{tab1},  the LieNet shows its superiority over the two baseline methods SO and SE. Besides, our LieNet with 3 blocks of RotMap and RotPooling layers achieves the best performance. For this dataset, we also study the performances of different block numbers in the LieNet architecture. As the number of frames in each sequence was fixed to 100 as mentioned before, adding more blocks to the LieNet will finally degenerate the temporal sequence into only 1 frame. In theory, this extreme case would result in the loss of the temporal resolution and thus undermine the performance of recognizing activities. In order to keep the balance between compact spatial feature learning and temporal information encoding, we therefore equip the LieNet with a limited number of blocks in different settings. Thus, we study 4 blocks at most for our LieNets. For the case of stacking 4 blocks, we find its performance (87.28\%) is lower than that of the 3-block case, which justifies the above argumentation. Nevertheless, as observed from Table \ref{tab1}, stacking some more RotMap/RotPooling blocks can improve the performance of the proposed LieNet.


\begin{table}
	\begin{center}
		\begin{tabular}{|l|c|}
			\hline
			Method & G3D-Gaming \\
			\hline\hline
			RBM+HMM \cite{nie2014capturing} & 86.40\% \\
			\hline\hline
			SE \cite{vemulapalli2014human} & 87.23\% \\
			SO \cite{vemulapallirolling} & 87.95\% \\			
			\hline
			\hline
			LieNet-0Block & 84.55\% \\
			LieNet-1Block &  85.16\% \\
			LieNet-2Blocks &  86.67\% \\
			LieNet-3Blocks & \textbf{89.10\%} \\
			\hline
		\end{tabular}
	\end{center}
	\caption{Recognition accuracies on the G3D-Gaming database.}
	\label{tab1}
\end{table}

\begin{figure}
	\begin{center}
		\includegraphics[width=1\linewidth]{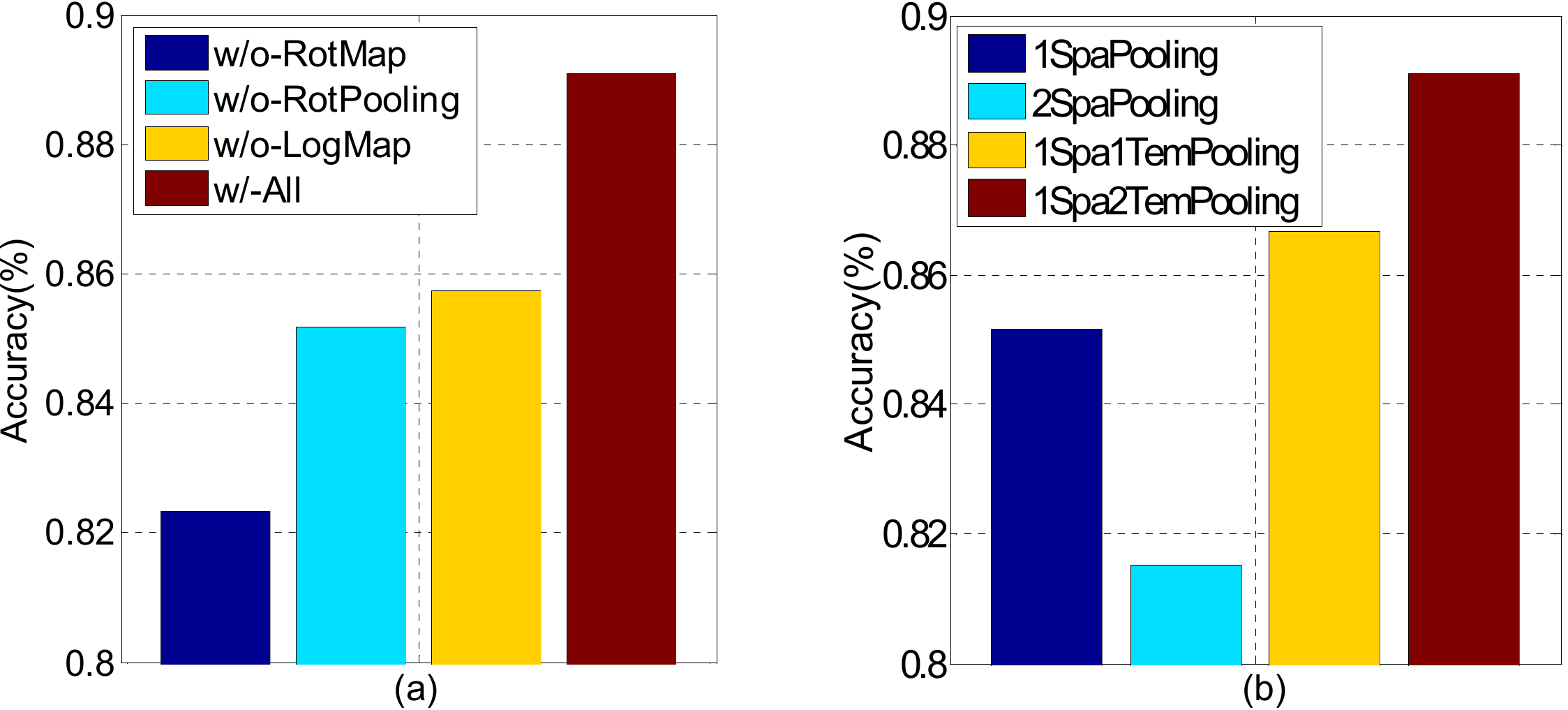}
		\caption{(a) Comparison of different LieNet configurations: without using RotMap layers (w/o-RotMap), w/o-RotPooling layers, w/o-LogMap layers and using all (w/-All) in LieNet-3Blocks for G3D-Gaming. (b) Comparison of different pooling schemes: using 1 spatial RotPooling layer (1SpaPooling, i.e., LieNet-1Block), 2 spatial RotPooling layers (2SpaPooling), 1SpaPooling+1 temporal RotPooling layer (1Spa1TemPooling, i.e., LieNet-2Blocks), 1SpaPooling+2TemPooling (1Spa2TemPooling, i.e., LieNet-3Blocks) for G3D-Gaming.}
		\label{Fig5}
	\end{center}
\end{figure}

\begin{figure}
	\begin{center}
		\includegraphics[width=1\linewidth]{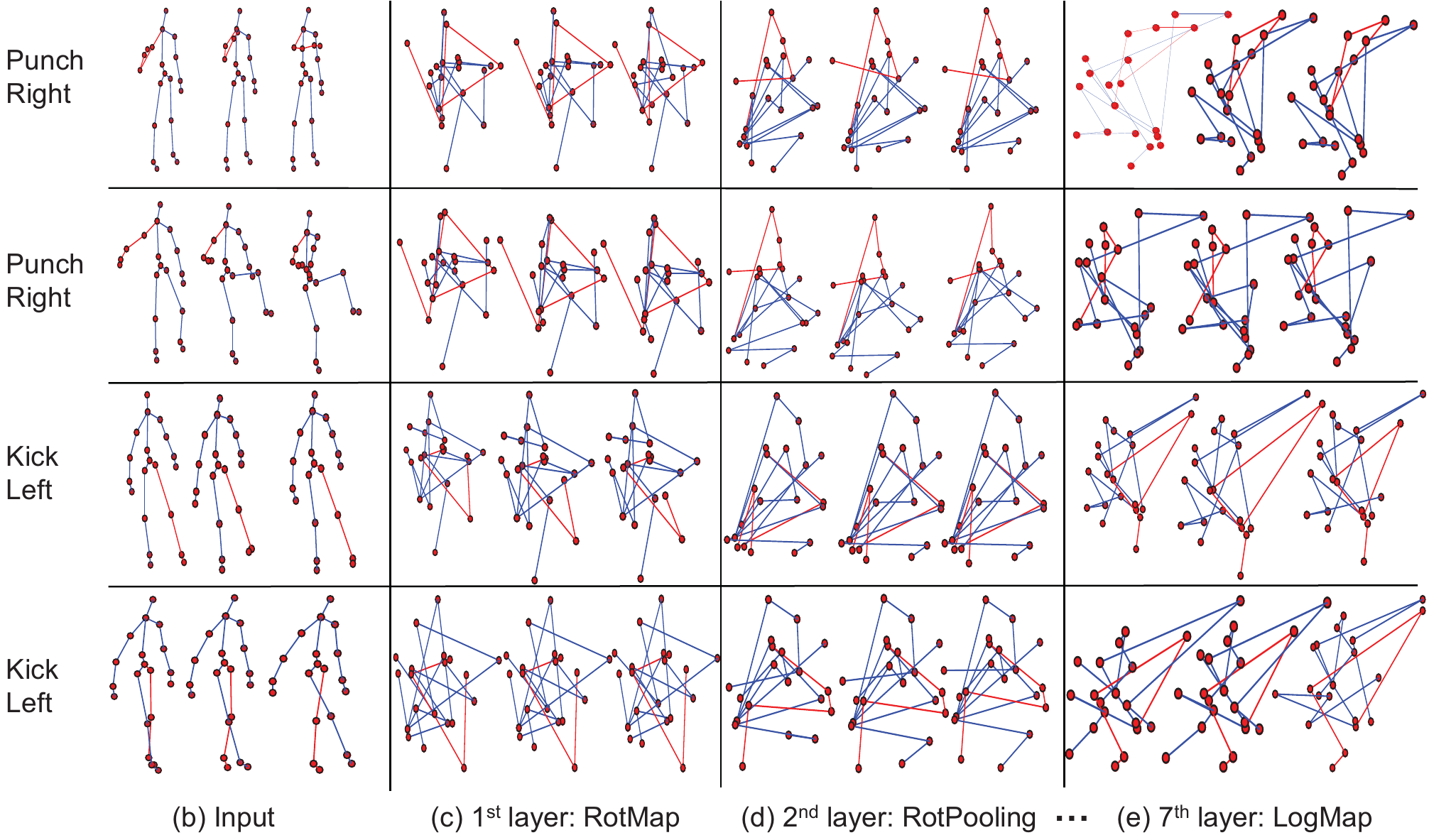}
		\caption{The example skeletons are reconstructed by the output rotation matrices of some representative LieNet layers for processing four action sequences from the G3D-Gaming dataset. The bones in red are the interesting ones for the action classes.}
		\label{Fig4}
	\end{center}
\end{figure}

In addition, we evaluate the performances of different LieNet configurations as shown in Fig.\ref{Fig5}. The left of Fig.\ref{Fig5} verifies the necessity of using RotMap, RotPooling and LogMap layers to improve the proposed LieNet-3Blocks. In addition, we also compare the LieNet with and without DTW. On this dataset, the performance (88.89\% vs. 89.10\%) of these two cases is approximately equal. Therefore, the benefit of using RotMap layers somehow shows it can take the role of DTW that solves the problem of speed variations. The right of Fig.\ref{Fig5} analyses the effectiveness of the 1SpaPooling case (i.e., Fig.\ref{Fig2}(a)$\rightarrow$(b)), and shows the decreasing performance behavior of the 2SpaPooling case (i.e., Fig.\ref{Fig2}(a)$\rightarrow$(b)$\rightarrow$(c)). Thus, we finally utilize 1SpaPooling and 2TemPooling (i.e., Fig.\ref{Fig2}(c)$\rightarrow$(d)) in the LieNet-3Blocks structure. Besides, we also study the behavior of adding a rectified linear unit (ReLU)-like layer (i.e., setting the matrix elements below a threshold $\epsilon=0.1$ to zero) on the top of the LogMap layer as presented before. Yet, the performance was worse (87.58\%) than without. Further, to validate the improvements are from the contribution of the RotMap and RotPooling layers rather than deeper architectures, we build a regular (LeNet-like) deep structure, i.e., LogMap$\rightarrow $2$\times$(FC$\rightarrow$MaxPooling)$\rightarrow$FC$\rightarrow$ReLU $\rightarrow$FC$\rightarrow$Softmax, that applies 8 regular layers on the concatenated output Euclidean forms of the LogMap layer. The step for MaxPooling is set to 4, and the sizes of different FC weights are set to $307800 \times 40000$, $10000 \times 4000$, $1000 \times 400$ and $400 \times 20$ respectively. The performance of this network is 85.49\%, which supports the validation.

For a better understanding of the proposed LieNet, we also visualize the output results of some representative layers. In particular, we roughly estimate the 3D location of each body bone, given the learned rotation matrix and the 3D coordinate of the beginning edge in the torso part. In Fig.\ref{Fig4}, we present the visualization of some layers for four action sequences, that belong to the classes of `punch right' and `kick left'. As shown in Fig.\ref{Fig4}, we observe that they yield meaningful semantic information layer by layer for specific classes. Specifically, the reconstructions from the first layer (RotMap) and the second layer (RotPooling) typically still mix some patterns specific for the action classes with some rather confusing ones. But, when arriving at the the seventh layer (LogMap), the patterns for specific motion classes become more discriminative.

\noindent \textbf{HDM05 dataset \cite{muller2007hdm}}. Following \cite{huang2016riemannian}, we conduct 10 random evaluations, each of which randomly selects half of the sequences for training and the rest for testing.

As listed in Table \ref{tab2}, besides to the two baseline methods SE and SO, we also study the SPDNet method \cite{huang2016riemannian} that has reached the best performance so far for this dataset. The large improvement of SE and SO over SPDNet suggests the effectiveness of the Lie group representations for the problem of skeleton-based action recognition. As the last experiment on the G3D-Gaming dataset, we also study the proposed LieNet with different numbers of blocks of RotMap and RotPooling layers. Note that since the length of each sequence in this database is fixed to 16 frames, as studied in the last evaluations, adding too much LieNet blocks will lead to the loss of the temporal resolution. Thus, we implemented the LieNet with 3 blocks at most for the dataset. As adding 3 blocks will generate 1 frame for each video, its performance (70.42\%) is not as promising as other cases. In contrast, as reported in Table \ref{tab2}, using more blocks (below 3 blocks) improves over using less blocks, and gets the state-of-the-art on the dataset, again showing its advantages over SE and SO shallow learning methods.

\begin{table}
	\begin{center}
		\begin{tabular}{|l|c|}
			\hline
			Method & HDM05 \\
			\hline\hline
			SPDNet \cite{huang2016riemannian} & 61.45\%$\pm$1.12\\
			\hline\hline
			SE \cite{vemulapalli2014human} & 70.26\%$\pm$2.89 \\
			SO \cite{vemulapallirolling}  & 71.31\%$\pm$3.21\\
			\hline
			\hline
			LieNet-0Block & 71.26\%$\pm$2.12 \\
			LieNet-1Block &  73.35\%$\pm$1.14\\
			LieNet-2Blocks &  \textbf{75.78\%$\pm$2.26} \\
			\hline
		\end{tabular}
	\end{center}
	\caption{Recognition accuracies on the HDM05 database.}
	\label{tab2}
\end{table}

\begin{table}
	\begin{center}
		\begin{tabular}{|l|c|c|}
			\hline
			Method & RGB+D-subject & RGB+D-view \\
			\hline\hline
			HBRNN \cite{du2015hierarchical} & 59.07\% & 63.97\% \\
			Deep RNN \cite{shahroudy2016ntu} & 56.29\% & 64.09\% \\
			Deep LSTM \cite{shahroudy2016ntu} & 60.69\% & 67.29\% \\	
			PA-LSTM \cite{shahroudy2016ntu} & 62.93\% & 70.27\% \\	
			ST-LSTM \cite{liu2016spatio} & \textbf{69.2}\% & \textbf{77.7\%} \\
			\hline\hline
			SE \cite{vemulapalli2014human} & 50.08\% & 52.76\%\\
			SO \cite{vemulapallirolling} & 52.13\% & 53.42\%\\		
			
			\hline
			\hline
			LieNet-0Block & 53.54\% & 54.78\%\\
			LieNet-1Block &  56.35\% & 60.14\%\\
			LieNet-2Blocks & 58.02\% & 62.52\%\\
			LieNet-3Blocks & 61.37\% & 66.95\%\\
			\hline
		\end{tabular}
	\end{center}
	\caption{Recognition accuracies for the cross-subject and cross-view evaluations on the NTU RGB+D database.}
	\label{tab3}
\end{table}

\noindent \textbf{NTU RGB+D dataset \cite{shahroudy2016ntu}}. This dataset has two standard testing protocols. One is cross-subject test, for which half of the subjects are used for training and the rest is for testing. The other one is cross-view test, for which two views are employed for training and the rest one is utilized for testing.

Since this dataset is large enough to train deep networks, recent works \cite{shahroudy2016ntu,liu2016spatio} studied typical Recurrent Neural Networks (deep RNN and deep LSTM) as well as two variants, i.e., part-aware (PA) and spatio-temporal (ST) versions of LSTM. The common advantage of these deep networks is to learn temporal information, and significantly outperform the Lie group representation learning methods SE and SO, which are good at learning spatial information, but are not deep learning models. In this paper, our LieNet fills the gap by showing the effectiveness of deep learning on the spatial representations. As shown in Table \ref{tab3}, our LieNet with more stacked blocks can significantly improve the two baseline methods SE and SO, which validates the effectiveness of the deep learning. By comparing with the state-of-the-art methods, our LieNet behaves better or equally well as most deep networks (e.g., deep RNN and deep LSTM) that exploit temporal information. The LieNet is still outperformed by the recently proposed PA-LSTM and ST-LSTM however, which jointly learn spatial and temporal features of moving skeletons. This is reasonable because the LieNet is mainly designed to learn the spatial features with only pooling temporal information.

\begin{figure}
	\begin{center}
		\includegraphics[width=0.85\linewidth]{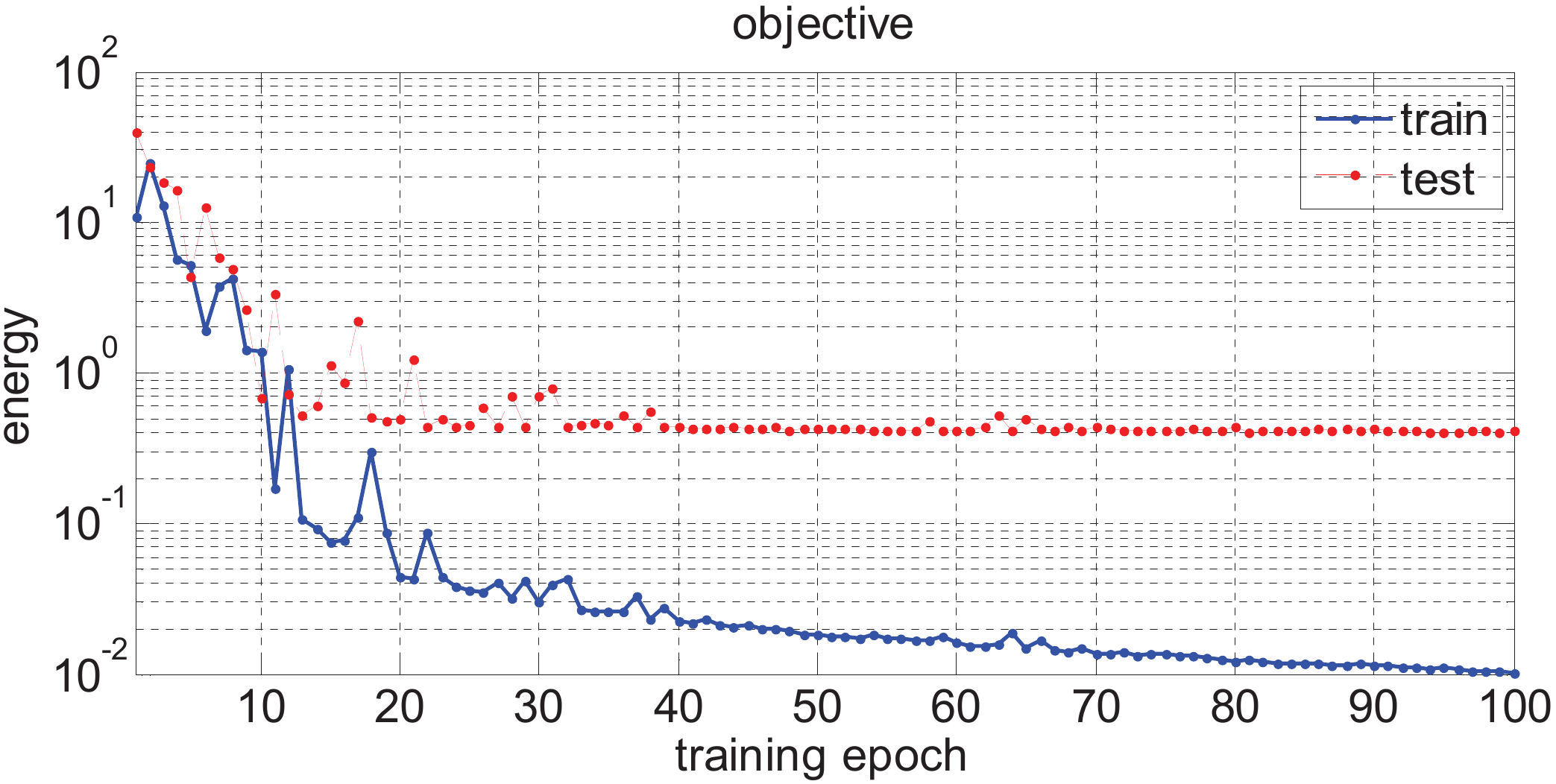}
		\caption{The convergence behavior of the proposed LieNet for the G3D-Gaming dataset.}
		\label{Fig6}
	\end{center}
\end{figure}

\noindent \textbf{Properties of LieNet training algorithm}. While the convergence of the used SGD algorithm on Riemannian manifolds has been studied well in \cite{bottou2010large, bonnabel2013stochastic} already, the convergence behavior (see Fig.\ref{Fig6}) of our LieNet training algorithm also demonstrates that it can converge to a stable solution after 100 epochs. In terms of the runtime, training LieNet-3Blocks takes about 6 minutes(m) per epoch on the G3D-gaming, and 514m per epoch on the NTU RGB+D. Training LieNet-2Blocks takes around 122m per epoch on the HDM05. In testing, the LieNet (i.e., the forward pass) takes about 3m, 4m and 86m respectively on G3D, HDM05 and NTU RGB+D. Note that, like usual network toolboxes, the current LieNet can be sped up a lot by implementing a GPU version. Regarding the memory requirement, LieNet-3Block-G3D, LieNet-2Block-HDM05 and LieNet-3Blocks-NTURGBD require around 1.2GB, 1.1GB and 1.4GB, respectively.


\section{Summary and Future Work}

We studied a deep network architecture in the domain of Lie group features, that is successful for skeleton-based action recognition. In order to handle the key issues of speed variation and high dimensionality of the Lie group features, we designed special mapping layers and pooling layers to process the resulting rotation matrices. In addition, we also exploited logarithm mapping layers to perform Riemannian computing on the representations, with which regular output layers are supplied in the new network structure. The final evaluations on three standard 3D action datasets not only demonstrated the effectiveness of the proposed network, but also compared its different configurations. Moreover, we also showed an interesting visualization for the network, which somewhat discloses its intrinsic mechanism.

As the proposed network is, to the best of our knowledge, the first attempt to perform deep learning on Lie groups for skeleton-based action recognition, there are quite a few open issues. For example, studying multiple rotation mappings per RotMap layer and exploiting a ReLU-like layer in the context of a Lie group network are worth paying attention to. Besides, building a deeper network, beginning from the raw 3D joint locations up to the Lie group features in an end-to-end learning manner, could be more effective. Last but not least, encouraged by the success of the deep spatio-temporal networks \cite{shahroudy2016ntu,liu2016spatio}, exploring the potential of the proposed network in the temporal setting would also be an interesting direction.

\noindent \textbf{Acknowledgments:}
This work is supported by EU Framework Seven project ReMeDi (grant 610902).


\end{document}